%% file: 00_main.tex
\documentclass[letterpaper, 10 pt, conference]{ieeeconf}
\IEEEoverridecommandlockouts
\usepackage{mathtools}
\usepackage{cite}
\usepackage[dvipsnames]{xcolor}
\usepackage{amsmath,amssymb,amsfonts}
\usepackage[font=small, labelfont=bf]{caption}
\usepackage[ruled, lined, linesnumbered, commentsnumbered, longend, noend]{algorithm2e}
\usepackage{hyperref}
\usepackage{algorithmic}
\usepackage{graphicx}
\usepackage{textcomp}
\usepackage{xcolor}
\usepackage{breqn}
\usepackage{gensymb}
\usepackage{cuted}
\usepackage{capt-of}
\usepackage{booktabs}
\usepackage{array}
\usepackage{diagbox}
\usepackage{pifont}
\usepackage{multirow}
\usepackage{makecell}
\usepackage{bm}

\newcommand{\cmark}{\textcolor{gray!150}{\scalebox{1.3}{\ding{51}}}}
\newcommand{\xmark}{\textcolor{gray!150}{\scalebox{1.3}{\ding{55}}}}
\newcommand{\start}{\textcolor{gray!110}{\ding{108}}}
\newcommand{\goal}{\textcolor{gray!110}{\ding{54}}}
\DeclareMathOperator*{\argmax}{arg\,max}
\DeclareMathOperator*{\argmin}{arg\,min}

\usepackage{fancyhdr} 
\def\BibTeX{{\rm B\kern-.05em{\sc i\kern-.025em b}\kern-.08em
    T\kern-.1667em\lower.7ex\hbox{E}\kern-.125emX}}
 
\newcommand{\argminE}{\mathop{\mathrm{argmin}}}

\definecolor{pink}{RGB}{255, 192, 203}
\definecolor{lightblue}{HTML}{77AADD}
\definecolor{lightorange}{HTML}{EE8866}
\definecolor{violet}{HTML}{C299DD}
\definecolor{tealgreen}{HTML}{44BB99}
\definecolor{lightyellow}{HTML}{D0B866}
\definecolor{apricot}{HTML}{DD9977}
\usepackage{siunitx}

\definecolor{darkgreen}{RGB}{83, 199, 34}

\title{\LARGE \bf \textit{FlowMP}: Learning Motion Fields for Robot Planning with \\Conditional Flow Matching}

\author{Khang Nguyen$^{1}$, An T. Le$^{2}$, Tien Pham$^{3}$, Manfred Huber$^{1}$, Jan Peters$^{2,4,5}$, and Minh Nhat Vu$^{6,7}$ 
\thanks{$^{1}$ Learning and Adaptive Robotics Lab, University of Texas at Arlington, USA. \href{mailto:khang.nguyen8@mavs.uta.edu}{\text{khang.nguyen8@mavs.uta.edu}}}
\thanks{$^{2}$Intelligent Autonomous Systems Lab, TU Darmstadt, Germany}
\thanks{$^{3}$Cognitive Robotics Lab, University of Manchester, UK} 
\thanks{$^{4}$German Research Center for AI (DFKI), SAIROL, Darmstadt, Germany} 
\thanks{$^{5}$Hessian.AI, Darmstadt, Germany}
\thanks{$^{6}$Automation \& Control Institute (ACIN), TU Wien, Vienna, Austria} 
\thanks{$^{7}$Austrian Institute of Technology (AIT) GmbH, Vienna, Austria}
}

\begin{document}

\maketitle
\thispagestyle{empty}
\pagestyle{empty}
\graphicspath{{./figures/}}

\begin{abstract}
    Prior flow matching methods in robotics have primarily learned \emph{velocity} fields to morph one distribution of trajectories into another. In this work, we extend flow matching to capture second-order trajectory dynamics, incorporating acceleration effects either explicitly in the model or implicitly through the learning objective. Unlike diffusion models, which rely on a noisy forward process and iterative denoising steps, flow matching trains a continuous transformation (flow) that directly maps a simple prior distribution to the target trajectory distribution without any denoising procedure. By modeling trajectories with second-order dynamics, our approach ensures that generated robot motions are smooth and physically executable, avoiding the jerky or dynamically infeasible trajectories that first-order models might produce. We empirically demonstrate that this second-order conditional flow matching yields superior performance on motion planning benchmarks, achieving smoother trajectories and higher success rates than baseline planners. These findings highlight the advantage of learning acceleration-aware motion fields, as our method outperforms existing motion planning methods in terms of trajectory quality and planning success.  
\end{abstract}

\input{01_introduction}
\input{02_related_work}
\input{03_methodology}
\input{04_evaluation}
\input{05_experiments}
\input{06_conclusions}

\bibliographystyle{IEEEtran}
\bibliography{IEEEabrv, references}

\end{document}

%% file: 01_introduction.tex
\section{Introduction}
\label{sec:intro}

Motion planning is a fundamental problem in robotics, and its applications range from autonomous navigation to robotic manipulation. As robots are deployed in increasingly complex and dynamic environments, generating collision-free, smooth, and dynamically feasible trajectories is crucial for reliable operation. Traditional motion planning methods can be broadly classified into sampling-based and optimization-based approaches. Optimization-based motion planners, such as CHOMP \cite{ratliff2009chomp} and sequential convex optimization methods \cite{schulman2014finding, mukadam2018continuous}, aim to refine an initial trajectory by minimizing a cost functional that typically encodes factors such as smoothness, collision avoidance, and dynamic feasibility. Although these methods can produce locally optimal solutions, their performance is highly dependent on the quality of the initial guess. Poor initialization, often a simple straight-line interpolation in configuration space \cite{ratliff2009chomp}, can lead to the convergence of suboptimal solutions or even complete failure, particularly in environments with narrow passages or cluttered obstacles.
Sampling-based methods like RRT-Connect \cite{kuffner2000rrt} guarantee probabilistic completeness by exploring the configuration space through random sampling. However, the trajectories produced by these methods are typically jerky and require additional smoothing or optimization steps \cite{leu2021efficient} to be suitable for execution in real robotic systems. 

\begin{figure}[t]
    \vspace{6pt}
    \centering
    \includegraphics[width=1.0\linewidth]{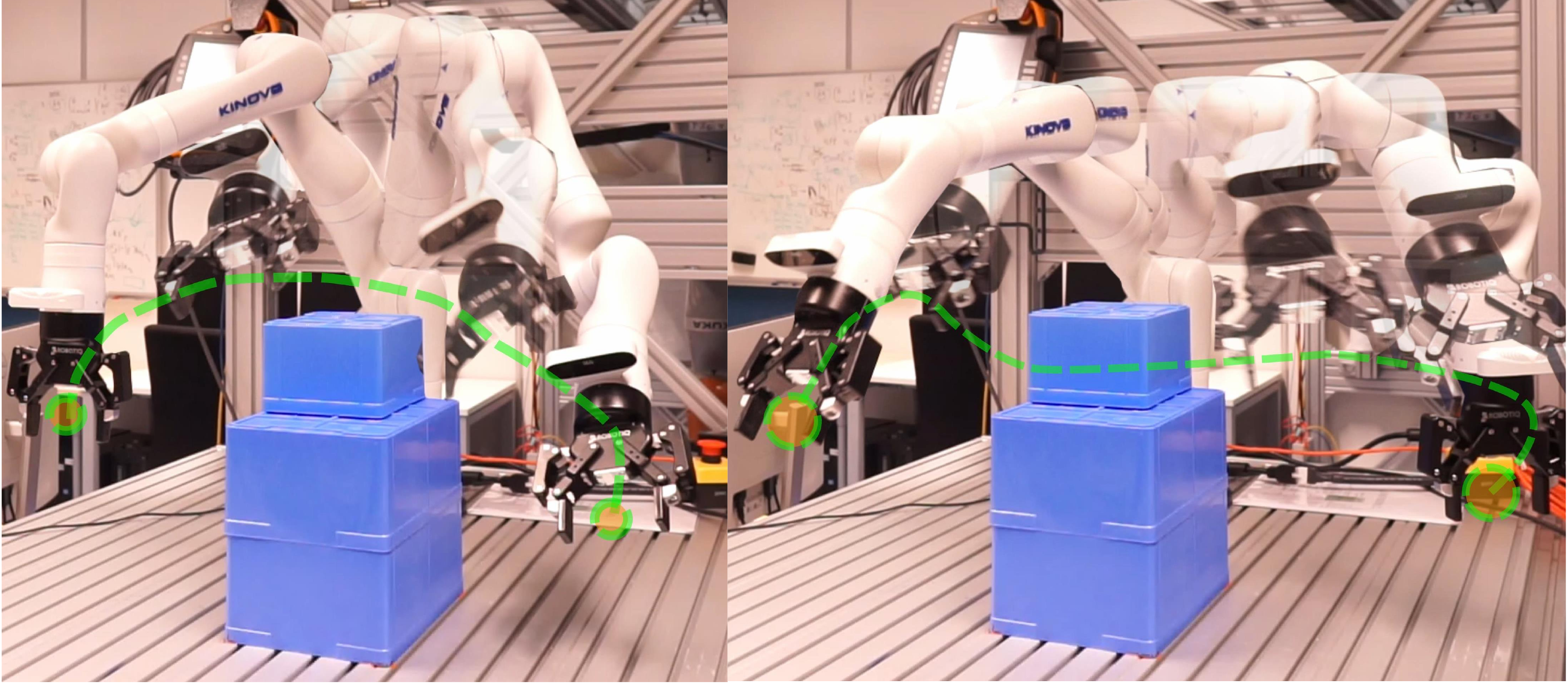}
    \caption{Executions of a smooth, dynamically feasible motion on a Kinova Gen3 manipulator. The same start and goal configurations can yield multiple valid solutions when sampling from our \textit{flow matching} framework, demonstrating its ability to capture different modes of the trajectory distribution.}
    \vspace{-18pt}
    \label{fig:overview_method}
\end{figure}

These limitations have motivated the search for more informed trajectory priors to guide planners toward high-quality solutions. Rather than sampling a new path from scratch, early work draws on a library of solution trajectories to warm-start the planner \cite{stolle2006policies, liu2009standing}. However, as the state dimension and database size grow, such memory-based methods suffer from the \emph{curse of dimensionality}. Subsequent approaches address this limitation by learning mappings from task parameters to trajectories. For instance, Mansard \textit{et al.} \cite{mansard2018using} used a neural network to predict an initial trajectory for a given task. In contrast, Lembono \textit{et al.} \cite{lembono2020memory}, and Power \textit{et al.} \cite{power2022variational} learned distributions of trajectories that allow sampling of diverse candidate paths.

Another line of research incorporates demonstrations of bias motion planning \cite{osa2018algorithmic}. Koert \textit{et al.} \cite{koert2016demonstration} introduced Demonstration-Based Trajectory Optimization (DEBATO), which encodes demonstrated motions as a probabilistic movement primitive (ProMP) and then optimizes the trajectory using a combination of collision avoidance and adherence to the ProMP prior through relative entropy policy search \cite{peters2010relative}. Rana \textit{et al.} \cite{rana2017clamp} proposed a Gaussian Process-based prior (CLAMP) to learn from demonstrations. In contrast, Urain \textit{et al.} \cite{urain2022learning} used an energy-based model to capture multimodal trajectory distributions. However, these approaches are generally unimodal or computationally intensive when handling complex, multimodal tasks.

Deep generative models have emerged as a promising way to represent complex trajectory distributions beyond the limitations of traditional parametric priors. Variational Autoencoders (VAE) \cite{kingma2014auto} and Generative Adversarial Networks (GAN) \cite{goodfellow2014generative} have been applied to learn data-driven priors but often struggle with training instabilities or mode collapse. Energy-based models offer another avenue \cite{lecun2006tutorial}, although they can be challenging to train and sample from effectively. More recently, diffusion models \cite{ho2020denoising, song2021score} have gained traction in capturing complex multimodal trajectory distributions. In particular, Carvalho \textit{et al.} \cite{carvalho2023motion} introduced Motion Planning Diffusion (MPD), which uses diffusion models to generate robot trajectories conditioned on task constraints. However, these typically operate by simulating iterative denoising steps and usually model only first-order dynamics (\textit{e.g.}, joint positions and velocities), which can result in trajectories lacking smooth acceleration profiles.

In this work, we propose to address these limitations by leveraging \emph{flow matching} as an alternative generative modeling technique for motion planning. Flow matching \cite{lipman2022flow, funk2024actionflow, zhang2024affordance} directly learns a continuous transformation—a motion field—that transports a simple prior distribution (\textit{e.g.}, Gaussian noise) to the distribution of expert trajectories without iterative denoising. Moreover, we extend the standard flow-matching approach by incorporating second-order trajectory dynamics, explicitly modeling acceleration alongside velocity to generate smoother and more dynamically feasible trajectories. Our contributions are twofold.
\begin{itemize}
    \item We introduce a conditional flow matching framework as a trajectory prior based on B-Spline representation to robot motion planning, offering a simulation-free, direct mapping from noise to feasible trajectories.
    \item We extend the flow matching approach to capture second-order dynamics, ensuring that the generated trajectories exhibit continuous velocity and acceleration profiles, critical for real-world robotic execution.
\end{itemize}

\begin{figure*}[t]    
    \centering
    \includegraphics[width=1.00\linewidth]{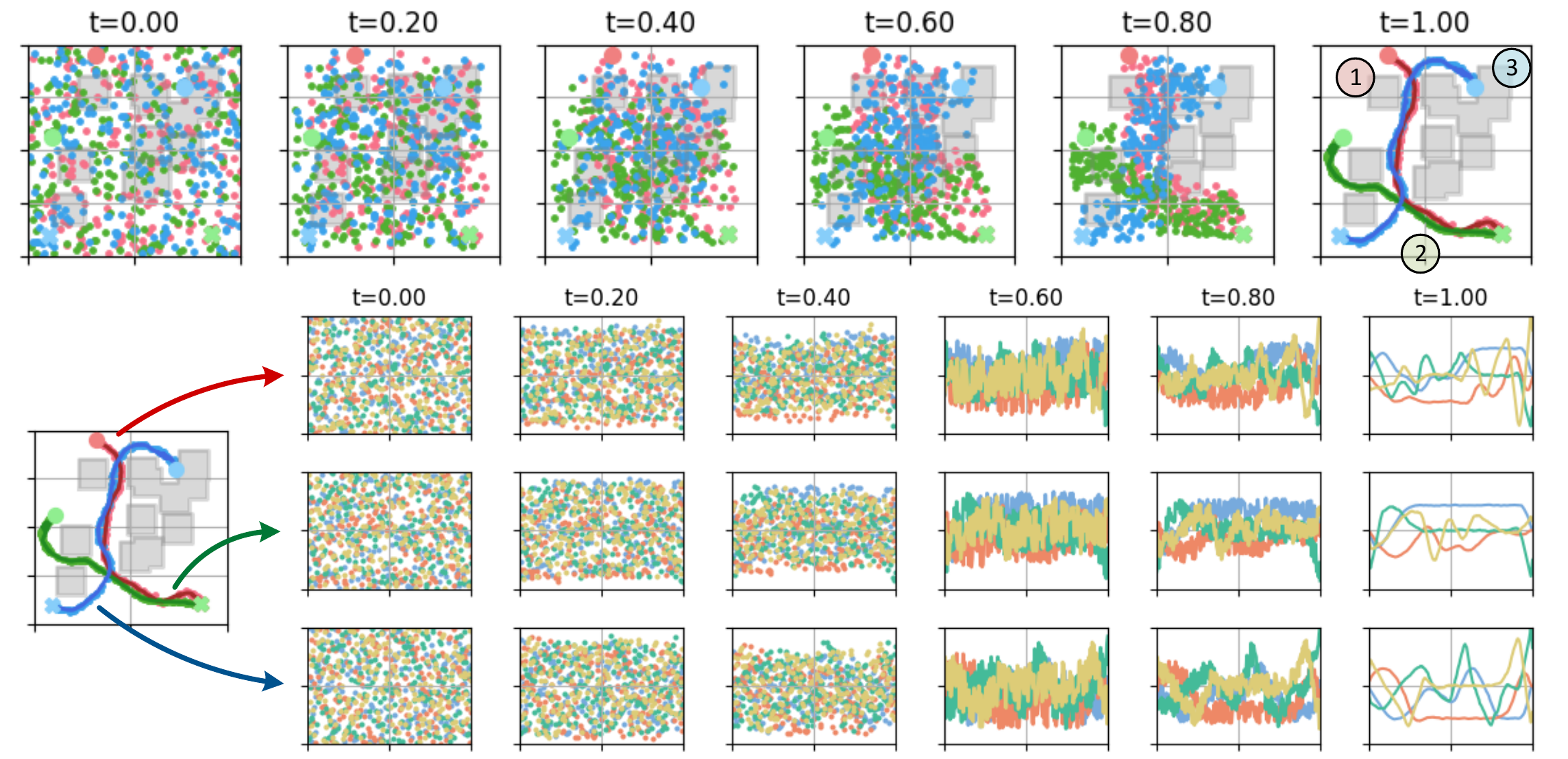}
    \vspace{-14pt}
    \caption{Motion planning at denoising steps along the time horizon $t$ from $0$ to $1$, inferred by the trained conditional motion field as outlined in the \texttt{generate_motion} function (Alg. \ref{alg:flow_motion_planning}). The paths correspond to different task objectives within the same \textcolor{Gray}{\textbf{maze environment}}: \textcolor{CarnationPink}{\textbf{Path 1}} and \textcolor{LimeGreen}{\textbf{Path 2}} start from different initial positions but share the same goal, while \textcolor{CornflowerBlue}{\textbf{Path 3}} has distinct start and goal points. The start and goal positions are indicated as \start~and~\goal, respectively, and are encoded as the same color of the paths. Their corresponding velocity (\textcolor{lightblue}{\textbf{blue}} and \textcolor{lightorange}{\textbf{orange}}) and acceleration (\textcolor{tealgreen}{\textbf{green}} and \textcolor{lightyellow}{\textbf{yellow}}) profiles of the simultaneously inferred paths are shown accordingly.}
    \vspace{-16pt}
    \label{fig:matching_steps_motion}
\end{figure*}

%% file: 02_related_work.tex
\section{Related Work}

\noindent\textbf{Warm-Starting Optimization-Based Motion Planning:} Classical trajectory optimizers require a good initial guess to converge to a feasible solution. Without informative priors, a straight-line interpolation between the start and goal is often used \cite{ratliff2009chomp}. Still, such initialization may lead to collisions or poor local minima \cite{schulman2014finding, leu2021efficient}. Early approaches addressed this by storing libraries of solution trajectories and retrieving the nearest neighbor for warm-starting \cite{stolle2006policies, liu2009standing}. However, these memory-based methods do not scale well with increasing state dimensions and dataset sizes. More recent techniques have focused on learning mappings from task parameters to trajectories. For instance, Mansard \textit{et al.} \cite{mansard2018using} proposed a neural network-based predictor, while methods by Lembono \textit{et al.} \cite{lembono2020memory} and Power and Berenson \cite{power2022variational} learned trajectory distributions that enable sampling of diverse candidates, thus improving the robustness of trajectory optimization. To handle trajectory multimodality, an energy-based model (EBM)~\cite{urain_2022_learning_implicit_priors} represented state density using an energy function and generates trajectories via stochastic optimization. However, EBM learning is challenging.

\noindent\textbf{Motion Planning with Diffusion Models:} Recent works have explored score-based and diffusion models in robotics as expressive multimodal generative priors for various tasks. In task planning, \cite{Kapelyukh2022-dall-e-bot} introduced \textsc{DALL-E-Bot} to generate goal scenes, while \cite{liu2022structdiffusion} proposed StructDiffusion for object arrangement. Diffusion models have also been applied to grasping~\cite{urain2022se3diffusion} and dexterous manipulation~\cite{DBLP:journals/corr/abs-2407-09899}. For behavior cloning, Diffusion Policy~\cite{DBLP:conf/rss/ChiFDXCBS23} and BESO~\cite{DBLP:conf/rss/ReussLJL23} modeled action distributions with diffusion models, forming the basis for advancements in perception, language conditioning, and collision avoidance~\cite{Ze2024DP3,DBLP:journals/corr/abs-2407-01812}. In trajectory planning, \cite{janner2022planning} introduced Diffuser, which models trajectories with a U-Net. Carvalho \textit{et al.} \cite{carvalho2023motion} introduced MPD that generates trajectories through iterative denoising, while its follow-up work \cite{carvalho2024motion} extended this framework to various environments and tasks. Despite their effectiveness, diffusion models require many iterations for trajectory generation and primarily focus on first-order dynamics, often neglecting smooth acceleration profiles. Other work~\cite{power2023sampling} used diffusion models for initializing constrained trajectory optimization. SafeDiffuser~\cite{DBLP:journals/corr/abs-2306-00148} applied control barrier functions for safety-critical tasks, and APEX~\cite{DBLP:journals/corr/abs-2404-02284} integrated diffusion with collision avoidance for bimanual manipulation.

\noindent\textbf{Flow Matching for Trajectory Generation:} Flow matching \cite{lipman2022flow, chen2024flow} has emerged as a promising alternative to diffusion models for generative modeling. Instead of relying on iterative denoising, flow matching directly learns a time-dependent vector field that transports a simple base distribution to the target distribution, generating high-quality samples in a few inference steps. In behavioral cloning, Funk \textit{et al.} \cite{funk2024actionflow} extended \cite{urain2022se3diffusion} by utilizing flow matching for efficient action generation. Braun \textit{et al.} \cite{braun2024riemannian} then generalized behavioral cloning flow matching to Riemannian manifold. In task planning, \cite{xu2024flow, zhang2024affordance} proposed a language-conditioned flow-matching model that maps visual observations or affordance to robot embodiments for task generation. 

In prior applications, flow matching has been used primarily to learn first-order trajectory distributions; however, in complex robotic systems, generating smooth and dynamically feasible trajectories requires capturing higher-order dynamics, such as acceleration. Our approach extends the standard flow-matching paradigm to incorporate second-order dynamics, resulting in a motion field that produces trajectories with continuous velocities and accelerations. This extension eliminates the need for iterative denoising and ensures that the generated trajectories are better suited for accurate robot execution.

%% file: 03_methodology.tex
\section{Flow Motion Planning}
\label{sec:methodology}

\subsection{Expert Via-Points Construction}
Trajectories can be represented in joint space as a weighted superposition of basis functions \cite{jankowski2022key, jankowski2023vp} as follows:
\begin{equation}
    \mathbf{q}(s) = \mathbf{\Phi}(s) \begin{bmatrix} \mathbf{q}_{\text{via}}^\top \\ \mathbf{w}_{\text{bc}}^\top \end{bmatrix} = \mathbf{\Phi}_{\text{via}}(s) \mathbf{q}_{\text{via}} + \mathbf{\Phi}_{\text{bc}}(s)\mathbf{w}_{\text{bc}},
    \label{eq:trajectory_representation}
\end{equation}
where $\mathbf{q}(s)$ is the trajectory at the phase $s = t/T \in \mathcal{S}$ with $T$ is the total movement duration, $\mathbf{\Phi}(s)$ represents the basis function matrix derived from splines, $\mathbf{q}_{\text{via}} = [\mathbf{q}_1^\top, \mathbf{q}_2^\top, \dots, \mathbf{q}_{n}^\top]^\top$ and $\mathbf{w}_{\text{bc}} = [\mathbf{q}_0^\top, \mathbf{q}'_0{}^\top, \mathbf{q}_T^\top, \mathbf{q}'_T{}^\top]^\top$ denote the $n$ via-points and boundary condition parameters, with $\mathbf{q}_0, \mathbf{q}_T$ are start and end positions and  $\mathbf{q}'_0, \mathbf{q}'_T$ are start and end velocities, coupling with $\mathbf{\Phi}_{\text{via}}(s)$ and $\mathbf{\Phi}_{\text{bc}}(s)$ represent the basis function matrices derived from corresponding via-points and boundary conditions in $\mathbf{\Phi}(s)$, respectively.

To find smooth and time-optimal trajectories going through optimal via-points $\mathbf{q}_{\text{via}}^{*}$, the cost function is minimized in the phase domain $\mathcal{S} \in [0, 1]$ (or $t = 0, 1, \dots, T$):
\begin{equation}
    \mathbf{q}_{\text{via}}^{*} = \argminE_{\mathbf{q}_{\text{via}}} \int_0^1 c\left(\mathbf{q}(s), \mathbf{\dot{q}}(s), \mathbf{\ddot{q}}(s)\right) ds,
    \label{eq:cost_minimizer}
\end{equation}
where $c(\cdot)$ is a task-specific cost function of collision avoidance, timing, and smoothness for kinodynamically valid and smooth trajectories, and $\mathbf{\dot{q}}(s)$, $\mathbf{\ddot{q}}(s)$ are derivatives of Eq. \ref{eq:trajectory_representation}.
\begin{equation*}
    \text{\small{$\mathbf{\dot{q}}(s) = \frac{1}{T} \mathbf{\Phi}'(s)\begin{bmatrix} \mathbf{q}_{\text{via}}^\top \\ \mathbf{w}_{\text{bc}}^\top \end{bmatrix} \quad \text{and} \quad \mathbf{\ddot{q}}(s) = \frac{1}{T^2} \mathbf{\Phi}''(s)\begin{bmatrix} \mathbf{q}_{\text{via}}^\top \\ \mathbf{w}_{\text{bc}}^\top \end{bmatrix}$}}.
    \label{eq:trajectory_derivatives}
\end{equation*}
The optimal via-points $\mathbf{q}_{\text{via}}$ for smooth and task-specific trajectories are optimized via the Covariance Matrix Adaptation Evolution Strategy (CMA-ES) algorithm \cite{hansen2016cma} by searching the via-point space and iteratively sampling candidate via-points from a Gaussian distribution $\mathcal{N}({\mu}_{\text{via}}, \mathbf{\Sigma}_{\text{via}})$, where ${\mu}_{\text{via}}$ is the mean and $\mathbf{\Sigma}_{\text{via}}$ is the covariance matrix. Each sampled set of via-points generates a trajectory through the parameterization in Eq. \ref{eq:trajectory_representation} and is evaluated using the cost function in Eq. \ref{eq:cost_minimizer}. CMA-ES then updates ${\mu}_{\text{via}}$ and $\mathbf{\Sigma}_{\text{via}}$ based on the evaluated costs, favoring candidate trajectories with lower costs and an optimal set of via-points $\mathbf{q}_{\text{via}}^{*}$.

We can hence generate B-Spline paths, $\tilde{\mathbf{q}}_{\text{via}}^{*}$, through optimal via-points $\mathbf{q}_{\text{via}}^{*} = [(x_i, y_i)]_{i = 1, 2, \dots, n}$:
\begin{equation}
    \scriptstyle
    \begin{aligned}
        \tilde{\mathbf{q}}_{\text{via}}^{*} &= \left[(\tilde{x}_i, \tilde{y}_i)\right] = \left[(x_i + \eta_{x_i}, y_i + \eta_{y_i})\right] \\
        &= \left[(x_i + W_{x}^{i} - W_{x}^{i-1}, y_i + W_{y}^{i} - W_{y}^{i-1})\right],
    \end{aligned}
    \label{eq:noisy_via_points}
\end{equation}
where $\eta_{x_i}, \eta_{y_i} \sim \mathcal{N}(0, \sigma^2)$, which can also be interpreted as discrete Brownian motions, $W_x$ and $W_y$, due to their incremental independence from $i-1$ to $i$.

Using noisy via-points $\tilde{\mathbf{q}}_{\text{via}}^{*}$, the B-splines are constructed with $\mathcal{C}_{3}^{i}(t)$ as cubic basis functions parametrized by time:
\begin{equation}
    \mathbf{q(}t) = \left[\mathbf{x}(t), \mathbf{y}(t)\right] = \left[\sum_{i=0}^{n} \tilde{x}_i \mathcal{C}_{3}^{i}(t), \sum_{i=0}^{n} \tilde{y}_i \mathcal{C}_{3}^{i}(t) \right].
\end{equation}
Respectively, the first and second derivatives of $\mathbf{q(}t)$ are collected, which are velocities and accelerations profiles. This trajectory construction, therefore, serves as the expert motion distribution between two random pairs of points.

\subsection{Conditional Motion Field}
To ensure the warm start for robot motions, we learn the distributions of positions, $\mathbf{Q}$, of expert paths along with their derivatives of positions, $\mathbf{\dot{Q}}$ and $\mathbf{\ddot{Q}}$ through conditional motion fields. We divide this motion field into velocity, acceleration, and jerk fields, corresponding to learning factors for position, velocity, and acceleration profiles of the trajectories, respectively. These conditional motion fields are learnable parameters that transition from noise to our target motion distributions via multiple probabilistic flow paths.

\setlength{\textfloatsep}{4pt}
\begin{algorithm}[t]
    \caption{Flow Motion Planning}
    \label{alg:flow_motion_planning}
    \begin{normalsize}
        \DontPrintSemicolon
        \SetKwInOut{KwIn}{Input}
        \SetKwInOut{KwOut}{Output}
        \SetKwFunction{FMainTrain}{motion\_field}
        \SetKwProg{Pn}{function}{}{}
        \KwIn{$ \pi_1 = \{ \mathbf{Q}, \mathbf{\dot{Q}}, \mathbf{\ddot{Q}} \} \coloneqq$ expert trajectories} 
        \KwOut{$\Upsilon^{\theta_1, \theta_2, \theta_3} = [\mathbf{u}^{\theta_1}, \mathbf{v}^{\theta_2}, \mathbf{w}^{\theta_3}] \coloneqq$ motion field}
        \Pn{\FMainTrain{$\pi_1$}}{
            \For{\texttt{in\_training}}{
                $t \gets U[0, 1]$ \\
                $\mathbf{q}_{1}, \mathbf{\dot{q}}_{1}, \mathbf{\ddot{q}}_{1} =  \pi_1[\texttt{batch\_size}]$ \\            
                $\varepsilon_{\mathbf{q}}, \varepsilon_{\mathbf{\dot{q}}}, \varepsilon_{\mathbf{\ddot{q}}} = \pi_0(\texttt{sampling\_size})$ \\
                $\mathbf{q}_{t} = \texttt{interp}(\mathbf{q}_1, \varepsilon_{\mathbf{q}}, \varepsilon_{\mathbf{\dot{q}}}, \varepsilon_{\mathbf{\ddot{q}}})$ \text{ (Eqs. \ref{eq:vel_interpolate}, \ref{eq:acc_interpolate}, \ref{eq:jerk_interpolate})} \\
                $\Upsilon \gets \texttt{get\_fields}(\mathbf{q}_1, \mathbf{\dot{q}}_{1}, \mathbf{\ddot{q}}_{1}, \varepsilon_{\mathbf{q}}, \varepsilon_{\mathbf{\dot{q}}}, \varepsilon_{\mathbf{\ddot{q}}})$ \\
                $\Upsilon_{t}^{\theta_1, \theta_2, \theta_3} = [\mathbf{u}_{t}^{\theta_1}, \mathbf{v}_{t}^{\theta_2}, \mathbf{w}_{t}^{\theta_3}] \gets f_{\theta_1, \theta_2, \theta_3}(\mathbf{q}_t, t)$ \\
                $\mathcal{L}(\theta) = || \Upsilon - \Upsilon_{t}^{\theta_1, \theta_2, \theta_3} ||_2^2$ \text{ (Eq. \ref{eq:total_loss})} \\
                $[\theta_1, \theta_2, \theta_3] \gets [\theta_1, \theta_2, \theta_3] + \nabla_{\theta_1, \theta_2, \theta_3} \mathcal{L}(\theta)$
            }
            \KwRet{$\Upsilon_{t}^{\theta_1, \theta_2, \theta_3}$}
        }
        \vspace{5pt}
        \SetKwFunction{FMainInfer}{generate\_motion}
        \KwIn{$ \pi_0 = \{ \varepsilon_{\mathbf{q}}, \varepsilon_{\mathbf{\dot{q}}}, \varepsilon_{\mathbf{\ddot{q}}} \} \coloneqq$ noise distributions \\
                $\delta \coloneqq$ step size in time;\\
                $\mathcal{O} \coloneqq $ task objective (start \& goal positions)} 
        \KwOut{$[\mathbf{q}, \mathbf{\dot{q}}, \mathbf{\ddot{q}}] \coloneqq$ robot motion}
        \Pn{\FMainInfer{$\pi_0$, $\delta$, $\mathcal{O}$ }}{
            $\mathbf{q}, \mathbf{\dot{q}}, \mathbf{\ddot{q}} = \varepsilon_{\mathbf{q}}, \varepsilon_{\mathbf{\dot{q}}}, \varepsilon_{\mathbf{\ddot{q}}} \gets \pi_0$ \\
            \For{$i \in [1, \ldots, [1 /\delta] ]$}{
                $\left[\mathbf{q}, \mathbf{\dot{q}}, \mathbf{\ddot{q}}\right] \gets \Upsilon_{t_i}^{\mathcal{O}}(\mathbf{q}, \mathbf{\dot{q}}, \mathbf{\ddot{q}}, \mathcal{O})$
            }
            \KwRet{$\mathbf{q}$, $\mathbf{\dot{q}}$, $\mathbf{\ddot{q}}$}
        }
    \end{normalsize}
\end{algorithm}

Denote $\pi_0 = \left\{ \varepsilon_{\mathbf{Q}}, \varepsilon_{\mathbf{\dot{Q}}}, \varepsilon_{\mathbf{\ddot{Q}}} \right\}$ and $\pi_1 = \left\{ \mathbf{Q}, \mathbf{\dot{Q}}, \mathbf{\ddot{Q}} \right\}$ as the noise and target motion distributions, respectively. We utilize rectified flow to shift from $\pi_0$ to $\pi_1$: velocity field maps $\varepsilon_{\mathbf{Q}}$ to $\mathbf{Q}$, acceleration field maps $\varepsilon_{\mathbf{\dot{Q}}}$ to $\mathbf{\dot{Q}}$, and jerk field maps from $\varepsilon_{\mathbf{\ddot{Q}}}$ to $\mathbf{\ddot{Q}}$, where $\varepsilon_{\mathbf{Q}}$, $\varepsilon_{\mathbf{\dot{Q}}}$, $\varepsilon_{\mathbf{\ddot{Q}}}$ represent the noises.

To generate the probabilistic flows, $p_{t}(\pi)_{0 \leq t \leq 1}$, from the source distribution, $\pi_0$, to the expert distribution, $\pi_1$, we find a motion field $\Upsilon$ of the time-dependent flow $\psi_{t}$:
\begin{equation}
    \frac{d}{dt} \psi_{t}(\pi_0) = \Upsilon_{t}(\psi_{t}(\pi)) 
    \label{eq:flow}
\end{equation}
where $\psi_{0}(\pi) = \pi \sim p_0$ and $\psi_{t}(\pi_0) = \pi_1 \sim p_{t}$.

\subsubsection{Conditional Velocity Field}
The time-differentiable interpolation from $\varepsilon_{\mathbf{Q}}$ to $\mathbf{Q}$ with the velocity field $\mathbf{u} = \mathbf{q}_{1} - \varepsilon_{\mathbf{q}}$ is formulated as:
\begin{equation}
    \mathbf{q}_{t} = (1 - t) \varepsilon_{\mathbf{q}} + t \mathbf{q}_{1}
    \label{eq:vel_interpolate}
\end{equation}

From Eq. \ref{eq:flow} and Eq. \ref{eq:vel_interpolate}, we obtain the conditional velocity field $\mathbf{u}_{t}(\mathbf{q} | \mathbf{q}_{1}) = (\mathbf{q}_{1} - \mathbf{q})/(1 - t)$. With Eq. \ref{eq:flow} and the conditional velocity field, we characterize the conditional probability flow $p_{t}(\mathbf{q})$ optimization governed by the velocity field loss $\mathcal{L}(\theta_1)$ with $t \sim U[0, 1]$, $\varepsilon_{\mathbf{q}} \sim \varepsilon_{\mathbf{Q}}$, and $\mathbf{q}_{1} \sim \mathbf{Q}$:
\begin{equation}
    \min_{\theta_1} \mathbb{E}_{\mathbf{q}_t, \mathbf{u}, t} \left[ f_{\theta_1}(\mathbf{q}_t, t) - \mathbf{u} \right] = \min \mathbb{E} \left( \mathbf{u}_t- \mathbf{u} \right)
    \label{eq:vel_field_loss}
\end{equation}

\subsubsection{Conditional Acceleration Field} 
Similar, with the acceleration field $\mathbf{v} = 2(\mathbf{q}_{1} - \varepsilon_{\mathbf{q}}) - 2 \varepsilon_{\mathbf{\dot{q}}}$ to learn the path velocity $\mathbf{\dot{q}}$, we formulate its time-differentiable interpolation:
\begin{equation}
    \mathbf{q}_{t} = (1 - t^2) \varepsilon_{\mathbf{q}} + (t - t^2) \varepsilon_{\dot{\mathbf{q}}} + t^2 \mathbf{q}_{1}
    \label{eq:acc_interpolate}
\end{equation}
and $\mathbf{v}_{t}(\mathbf{\dot{q}} | \mathbf{\dot{q}}_{1}) = (\mathbf{\dot{q}}_{1} - \mathbf{\dot{q}})/(1 - t)$ as the conditional acceleration field, inducing the acceleration field loss $\mathcal{L}(\theta_2)$:
\begin{equation}
    \min_{\theta_2} \mathbb{E}_{\mathbf{q}_t, \mathbf{v}, t} \left[ f_{\theta_2}(\mathbf{q}_t, t) - \mathbf{v} \right] = \min \mathbb{E} \left( \mathbf{v}_t- \mathbf{v} \right)
    \label{eq:acc_field_loss}
\end{equation}

\subsubsection{Conditional Jerk Field} 
With the jerk field $\mathbf{w} = 6(\mathbf{q}_1 - \varepsilon_{\mathbf{q}}) - 6\varepsilon_{\mathbf{\dot{q}}} - 3\varepsilon_{\mathbf{\ddot{q}}}$ that learns the path acceleration is derived, the time-differentiable interpolation is, therefore, given as:
\begin{equation}
    \mathbf{q}_{t} = (1 - t^3) \varepsilon_{\mathbf{q}} + (t -t^3) \varepsilon_{\mathbf{\dot{q}}} + \frac{t^2 - t^3}{2} \varepsilon_{\mathbf{\ddot{q}}} + t^3 \mathbf{q}_{1}
    \label{eq:jerk_interpolate}
\end{equation}
coupling with $\mathbf{w}_{t}(\mathbf{\ddot{q}} | \mathbf{\ddot{q}}_{1}) = (\mathbf{\ddot{q}}_{1} - \mathbf{\ddot{q}})/(1 - t)$ as the condition jerk field, resulting in the jerk field loss $\mathcal{L}(\theta_3)$:
\begin{equation}
    \min_{\theta_3} \mathbb{E}_{\mathbf{q}_t, \mathbf{w}, t} \left[ f_{\theta_3}(\mathbf{q}_t, t) - \mathbf{w} \right] = \min \mathbb{E} \left( \mathbf{w}_t- \mathbf{w} \right)
    \label{eq:jerk_field_loss}
\end{equation}

Leveraging Eq. \ref{eq:vel_field_loss}, Eq. \ref{eq:acc_field_loss}, and Eq. \ref{eq:jerk_field_loss}, we simultaneously train velocity, acceleration, and jerk fields, so-called motion field, $\Upsilon = \left\{ \mathbf{u}, \mathbf{v}, \mathbf{w} \right\}$ as depicted in Alg. \ref{alg:flow_motion_planning}, bringing noises $\pi_0$ to expert distributions $\pi_1$, including position, velocity, and acceleration profiles, via the total motion field loss:
\begin{equation}
    \mathcal{L}\left( \theta \right) =  \mathcal{L}(\theta_1) + \mathcal{L}(\theta_2) + \mathcal{L}(\theta_3) = || \Upsilon - \Upsilon_{t}^{\theta_1, \theta_2, \theta_3} ||_2^2
    \label{eq:total_loss}
\end{equation}

\subsection{Motion Field Inference}
With the trained motion field prior distribution $\pi_1$ in an environment, we can sample the field from the posterior $p(\Upsilon | \mathcal{O})$ given a task objective $\mathcal{O}$. Let $p(\mathcal{O} | \Upsilon)$ as the objective likelihood, we have:
\begin{equation}
    p(\Upsilon | \mathcal{O}) \propto p(\mathcal{O} | \Upsilon) \pi_1(\Upsilon)^{\lambda_{\text{prior}}},
    \label{eq:posterior_sampling}
\end{equation}
assumming the likelihood factorizes as~\cite{urain_2022_learning_implicit_priors} with $\lambda_j > 0$:
\begin{align}
    p(\mathcal{O} | \Upsilon) \propto \prod p_j(\mathcal{O}_j | \Upsilon)^{\lambda_j}.
    \label{eq:objective_likelihood}
\end{align}
Assuming the likelihood having the exponential form ${p_j(\mathcal{O}_j |  \Upsilon) \propto \exp(-C_j(\Upsilon))}$, we perform Maximum-a-Posteriori (MAP) on the trajectory posterior:
\begin{subequations}
    \small
    \begin{align}
        & \argmax_{\Upsilon} \log  p(\mathcal{O} | \Upsilon) \pi_1(\Upsilon)^{\lambda_{\text{prior}}}  \label{eq:mp_objective_1} \\
        &=  \argmax_{\Upsilon} \sum_{j} \log \exp(-C_j(\Upsilon))^{\lambda_j} + \log  \pi_1(\Upsilon)^{\lambda_{\text{prior}}} \label{eq:mp_objective_2} \\
        &=  \argmin_{\Upsilon}  \sum_{j} \lambda_j C_j(\Upsilon) - \lambda_{\text{prior}} \log \pi_1(\Upsilon),\label{eq:mp_objective_3}
    \end{align}
    \label{eq:mp_objective}
\end{subequations}
with the gradient derived from $i^{\mathrm{th}}$ inference step of Eq. \ref{eq:objective_likelihood}:
\begin{align}
    \mathbf{g}_{t_i} & = \nabla_{\Upsilon_{t_{i-1}}} \log p(\mathcal{O} | \Upsilon_{t_{i-1}}) \\
    &= - \sum_{j} \lambda_j \nabla_{\Upsilon_{t_{i-1}}} C_j(\Upsilon_{t_{i-1}}).
\end{align}

\begin{figure*}[t]    
    \centering
    \includegraphics[width=1.00\linewidth]{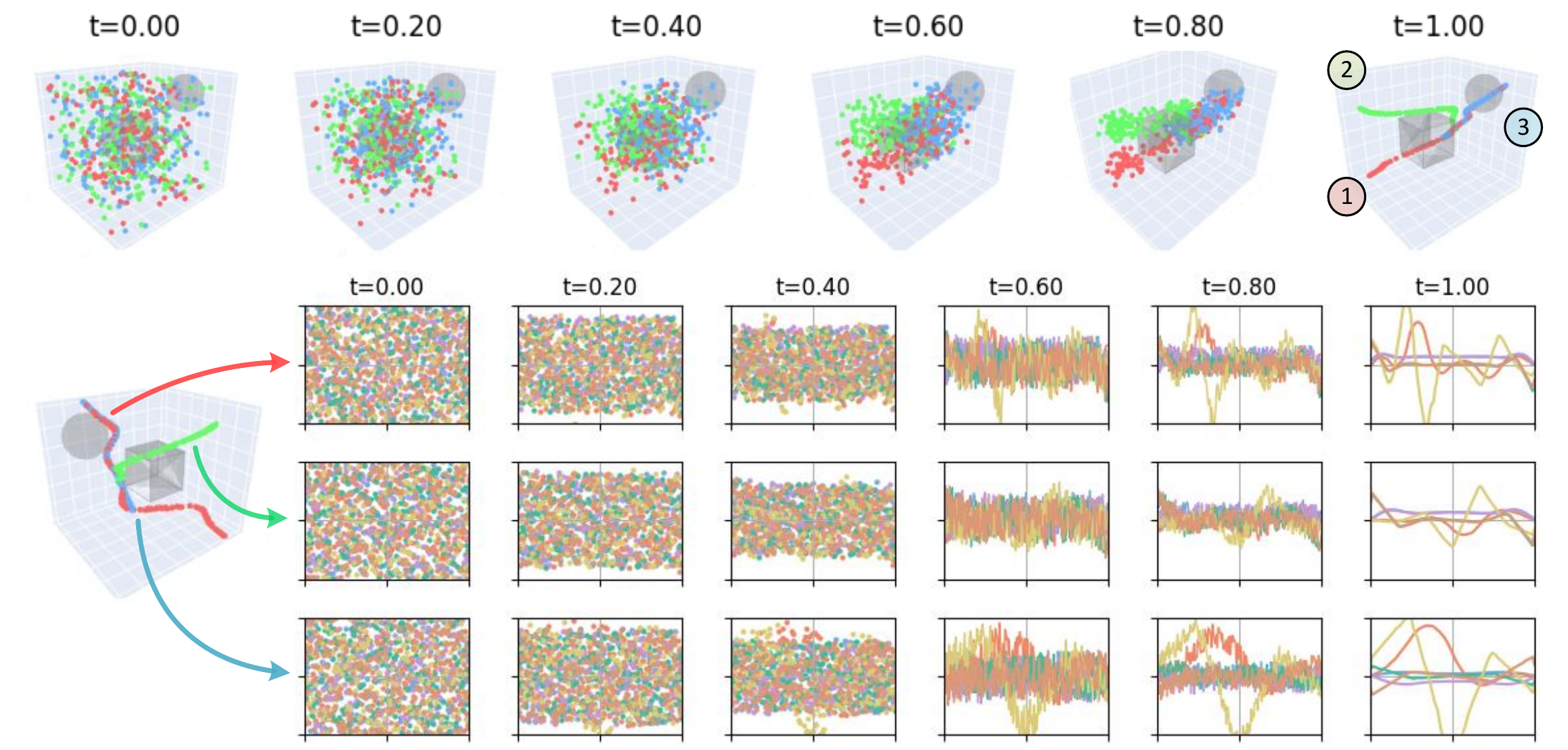}
    \vspace{-14pt}
    \caption{The conditional motion field is extended to 3D space, where collision-free paths are denoised along the time horizon $t$ from $0$ to $1$ within a \textcolor{Gray}{\textbf{obstacle map}} from the initial noises covering the workspace. The model is trained on expert motion distributions with various task objectives in the pre-defined 3D space using Alg. \ref{alg:flow_motion_planning}. Three example motions are sampled with the \texttt{generate_motion} function, where \textcolor{Red}{\textbf{Path 1}} and \textcolor{Turquoise}{\textbf{Path 3}} start from different initial positions but go to the same goal, whereas \textcolor{LimeGreen}{\textbf{Path 2}} follows a distinct trajectory with unique start and goal points. Velocity (\textcolor{lightblue}{\textbf{blue}}, \textcolor{lightorange}{\textbf{orange}}, and \textcolor{violet}{\textbf{violet}}) and acceleration (\textcolor{tealgreen}{\textbf{green}}, \textcolor{lightyellow}{\textbf{yellow}}, and \textcolor{apricot}{\textbf{apricot}}) profiles on three axes are sampled from their respective noise distributions. At $t = 1$, the trajectories and their derivatives are smooth.}
    \vspace{-15pt}
    \label{fig:matching_steps_motion_3d}
\end{figure*}

Given that the expert prior distribution $\pi_1$ provides sufficient coverage of the objective distribution $\mathcal{O}$, and the learned motion field $\Upsilon_{t}^{\theta_1, \theta_2, \theta_3}$ is parameterized by trainable parameters $(\theta_1, \theta_2, \theta_3)$ bridging $\pi_0$ and $\pi_1$, it follows that the learned motion field $\Upsilon$ acts as a transport operator mapping any sub-distribution of $\pi_0$ to a corresponding sub-distribution of $\pi_1$. In this case, the likelihood acts as an out-of-distribution guidance, straightforwardly leading to the posterior flow $\Upsilon_{t_i}^{\mathcal{O}} = \boldsymbol{g}_{t_i} + \lambda_{\text{prior}}\Upsilon_{t_i}^{\theta_1, \theta_2, \theta_3}$. Consequently, this guarantees that the optimization process preserves the statistical consistency required for task success, reinforcing the theoretical validity of the learned motion dynamics. In particular, the prior sampling mechanism (Eq. \ref{eq:posterior_sampling}) can be formulated as a fourth-order Runge-Kutta integration that propagates the motion field over the specified time horizon:
\begin{align}
    &k_1 = f(t_i, \Upsilon_{t_{i-1}}),\,k_2 = f\left(t_i + \frac{\delta}{2}, \Upsilon_{t_{i-1}} + \frac{\delta}{2} k_1\right) \nonumber \\
    &k_3 = f\left(t_i + \frac{\delta}{2}, \Upsilon_{t_{i-1}} + \frac{\delta}{2} k_2\right),\,k_4 = f(t_i + \delta, \Upsilon_{t_{i-1}} + \delta k_3) \nonumber\\
    &\Upsilon_{t_i}^{\theta_1, \theta_2, \theta_3} = \Upsilon_{t_{i-1}} + \frac{\delta}{6} \left(k_1 + 2k_2 + 2k_3 + k_4\right) \label{eq:rk4_final}
\end{align}
where $f = f_{\theta_1, \theta_2, \theta_3}$ represents our trained model leading $\pi_0$ to $\pi_1$, $t_{i}$ is timestamps from $0$ to $1$, and $\delta$ is the time step, which is equivalent to the underlying diffusion step in Eq. \ref{eq:posterior_sampling} in this context of motion field.

\begin{table}[t]
    \centering
    \begin{tabular}{r ccc}
        \toprule
        \multirow{3}{*}{\diagbox{Planner}{Length}} & \multicolumn{3}{c}{\makecell{2D Environment (\texttt{RobotPointMass})}} \\
        \cmidrule(lr){2-4}
        & $L = 64$ & $L = 128$ & $L = 256$ \\
        \midrule \midrule
        Stoch-GPMP \cite{urain2022learning} & \cmark & \cmark & \cmark \\
        \midrule
        MPD (DP) \cite{carvalho2023motion} & \xmark & \xmark & \xmark \\
        MPD (DG) \cite{carvalho2023motion} & \cmark & \xmark & \xmark \\
        MPD (full) \cite{carvalho2023motion} & \cmark & \xmark & \xmark \\
        \cmidrule[0.5pt]{1-4}
        FlowMP (ours) & \cmark & \cmark & \cmark \\
        \bottomrule
    \end{tabular}
    \caption{Planning feasibility of FlowMP, Stoch-GPMP \cite{urain_2022_learning_implicit_priors}, and MPD \cite{carvalho2023motion} in \texttt{RobotPointMass} environments.}
    \label{tab:planning_feasibility_2d}
\end{table}

Alg. \ref{alg:flow_motion_planning} recaps the training and inference processes. A sampling example is shown under motion fields in Fig. \ref{fig:matching_steps_motion}, resulting in smooth trajectories, velocity, and acceleration profiles with different task objectives in a maze environment. Fig. \ref{fig:matching_steps_motion_3d} further illustrates the extension of a conditional motion field in 3D space, where initially scattered samples are progressively refined into structured, collision-free paths as time progresses from $t = 0$ to $t = 1$. The top row shows the denoising process, where samples converge towards feasible trajectories. Meanwhile, the bottom row presents velocity and acceleration profiles, sampled from their respective noise distributions, demonstrating how different motion objectives shape the resulting kinematics. Both Fig. \ref{fig:matching_steps_motion} and Fig. \ref{fig:matching_steps_motion_3d} demonstrate denoised motions with a sampling size of 200.

\begin{table}[t]
    \centering
    \begin{tabular}{r ccc}
        \toprule
        \multirow{3}{*}{\diagbox{Planner}{Length}} & \multicolumn{3}{c}{\makecell{3D Environment (\texttt{PandaSphere})}} \\
        \cmidrule(lr){2-4}
        & $L = 64$ & $L = 128$ & $L = 256$ \\
        \midrule \midrule
        Stoch-GPMP \cite{urain2022learning} & \cmark & \cmark & \cmark \\
        \midrule
        MPD (DP) \cite{carvalho2023motion} & \cmark & \xmark & \xmark \\
        MPD (DG) \cite{carvalho2023motion} & \cmark & \xmark & \xmark \\
        MPD (full) \cite{carvalho2023motion} & \cmark & \xmark & \xmark \\
        \cmidrule[0.5pt]{1-4}
        FlowMP (ours) & \cmark & \cmark & \cmark \\
        \bottomrule
    \end{tabular}
    \caption{Planning feasibility of FlowMP, Stoch-GPMP \cite{urain_2022_learning_implicit_priors}, and MPD \cite{carvalho2023motion} in \texttt{PandaSphere} environment.}
    \label{tab:planning_feasibility_3d}
\end{table}

\begin{table*}[t]
    \vspace{5pt}
    \centering
    \caption{Inference time (\textit{in seconds}) of FlowMP compared to Stoch-GPMP and MPD. It is noted that the inference times of MPD with varied-length paths are still measured for reference purposes despite their planning infeasibility in Table \ref{tab:planning_feasibility_2d} and Table \ref{tab:planning_feasibility_3d}.}
    \label{tab:inference_time}
    \vspace{-2pt}
    \begin{tabular}{r ccc ccc}
        \toprule
        \multirow{2}{*}{\diagbox{Planner}{Length}} & \multicolumn{3}{c}{2D Environment (\texttt{RobotPointMass})} & \multicolumn{3}{c}{3D Environment (\texttt{PandaSphere})} \\
        \cmidrule(lr){2-4} \cmidrule(lr){5-7}
        & $L = 64$ & $L = 128$ & $L = 256$ & $L = 64$ & $L = 128$ & $L = 256$ \\
        \midrule \midrule
        Stoch-GPMP \cite{urain2022learning} & 0.3335 $\pm$ 0.0117 & 0.6896 $\pm$ 0.01694 & 2.1184 $\pm$ 0.0225 & 0.6114 $\pm$ 0.0050 & 0.8906 $\pm$ 0.0076 & 2.0008 $\pm$ 0.0196 \\
        \midrule
        MPD (DP) \cite{carvalho2023motion} & 0.1034 $\pm$ 0.0018 & 0.1018 $\pm$ 0.0019 & 0.1033 $\pm$ 0.0008 & 0.1406 $\pm$ 0.0025 & 0.1370 $\pm$ 0.0032 & 0.1374 $\pm$ 0.0032 \\
        MPD (DG) \cite{carvalho2023motion} & 0.2553 $\pm$ 0.0046 & 0.2584 $\pm$ 0.0050 & 0.2636 $\pm$ 0.0069 & 0.1396 $\pm$ 0.0024 & 0.1364 $\pm$ 0.0016 & 0.1367 $\pm$ 0.0020 \\
        MPD (full) \cite{carvalho2023motion} & 0.2556 $\pm$ 0.0034 & 0.2630 $\pm$ 0.0032 & 0.2647 $\pm$ 0.0034 & 5.1689 $\pm$ 0.1818 & 5.1091 $\pm$ 0.1602 & 5.2963 $\pm$ 0.1270 \\
        \cmidrule[0.5pt]{1-7}
        FlowMP (ours) &\textbf{0.0937 $\pm$ 0.0019} & \textbf{0.0941 $\pm$ 0.0015} & \textbf{0.0989 $\pm$ 0.0070} & \textbf{0.1302 $\pm$ 0.0059} & \textbf{0.1312 $\pm$ 0.0041} & \textbf{0.1341 $\pm$ 0.0077} \\
        \bottomrule
    \end{tabular}
    \vspace{-14pt}
\end{table*}

%% file: 04_evaluation.tex
\section{Evaluations \& Experiments}
\label{sec:evaluations}

We evaluate FlowMP against other baseline motion planners, including Stoch-GPMP \cite{urain_2022_learning_implicit_priors} and MPD \cite{carvalho2023motion} in terms of planning feasibility, trajectory smoothness, integration error, and inference time on the NVIDIA RTX 4070 GPU. All methods are made to generate a batch of $25$ trajectories during inference. Each runs with different trajectory lengths: $64$, $128$, and $256$. Stoch-GPMP runs with 700 iterations with a $\sigma_{GP}$ of $0.1$ for 2D environments and a $\sigma_{GP}$ of $0.0007$ for 3D environments, MPD infers the paths with three of their variants: diffusion prior (DP), diffusion prior then guide (DG), and their original approach. Both Stoch-GPMP and MPD only generate trajectories with velocities. In contrast, FlowMP additionally infers acceleration profiles alongside positions and velocities. For MPD and FlowMP, we run them at the same denoising steps of $30$ for comparison fairness.

\subsection{Planning Feasibilty}
First, we experiment with FlowMP, Stoch-GPMP, and MPD on \texttt{RobotPointMass} and \texttt{PandaSphere} environments to see their planning feasibility with varied-length paths. Table \ref{tab:planning_feasibility_2d} and Table \ref{tab:planning_feasibility_3d} show that Stoch-GPMP and FlowMP succeed in planning trajectories with diverse lengths. However, MPD raises scalability concerns as it collapses the trajectories with lengths of more than 64. This experiment proves that FlowMP is more scalable than MPD as a diffusion-based global planner while avoiding the myoptic failures of local planning-based methods. 

\begin{table}[t]
    \vspace{5pt}
    \centering
    \begin{tabular}{r cc}
        \toprule
        \diagbox{Planner}{Env} & \makecell{2D Environment \\ (\texttt{RobotPointMass})} & \makecell{3D Environment \\ (\texttt{PandaSphere})} \\
        \midrule \midrule
        Stoch-GPMP \cite{urain2022learning} & \textbf{0.0725} & 2.3713 \\
        \midrule
        MPD (DP) \cite{carvalho2023motion} & -- & 1.5638 \\
        MPD (DG) \cite{carvalho2023motion} & 0.1755 & 1.2415 \\
        MPD (full) \cite{carvalho2023motion} & 0.1268 & 0.2572 \\
        \cmidrule[0.5pt]{1-3}
        FlowMP (Ours) & 0.0846 & \textbf{0.1192} \\
        \bottomrule
    \end{tabular}
    \caption{Comparison of trajectory smoothness of paths with \textit{length of 64} generated by Stoch-GPMP, MPD, FlowMP.}
    \label{tab:trajectory_smoothness}
\end{table}

\subsection{Trajectory Smoothness}
Next, with generated valid collision-free trajectories, we evaluate their smoothness to guarantee that a robot can physically execute the motion while maintaining control feasibility. We simply compute the sum of point-to-point slopes from one end to another of a best generated trajectory. As shown in Table \ref{tab:trajectory_smoothness}, the results show that our trajectories are smoother compared to MPD, achieving second-best in \texttt{RobotPointMass} environment and smoothest in \texttt{PandaSphere} environment.

\begin{table}[t]
    \vspace{5pt}
    \centering
    \begin{tabular}{r cc}
        \toprule
        \multirow{3}{*}{\diagbox{Planner}{Env}} & \makecell{2D Environment \\ (\texttt{RobotPointMass})} & \makecell{3D Environment \\ (\texttt{PandaSphere})} \\
        & $(x, y)$ & $(x, y, z)$ \\
        \midrule \midrule
        Stoch-GPMP \cite{urain2022learning} & \textbf{(0.036, 0.009)} & \textbf{(0.205, 0.201, 0.199)}  \\
        \midrule
        MPD (DP) \cite{carvalho2023motion} & -- & (0.903, 0.096, 0.033) \\
        MPD (DG) \cite{carvalho2023motion} & (0.058, 0.018) & (0.920, 0.405, 0.053) \\
        MPD (full) \cite{carvalho2023motion} & (0.068, 0.021) & (0.315, 0.504, 0.408) \\
        \cmidrule[0.5pt]{1-3}
        FlowMP (Ours) & (0.043, 0.016) & (0.426, 0.321, 0.285) \\
        \bottomrule
    \end{tabular}
    \caption{Comparison of integration error along axes of the paths with \textit{length of 64} generated by Stoch-GPMP, MPD, FlowMP.}
    \label{tab:integration_error}
\end{table}

\subsection{Integration Error}
Furthermore, we utilize the Romberg quadrature method \cite{romberg1955vereinfachte} to quantify the integration error between the start and end points along dimensional axes of the best generated trajectory ($L = 64$) for all baselines and FlowMP, as shown in Table \ref{tab:integration_error}. This metric depicts that FlowMP achieves lower integrative error than MPD in \texttt{RobotPointMass} and \texttt{PandaSphere} environments. Meanwhile, Stoch-GPMP has the least errors in both cases.

\subsection{Inference Time}
Lastly, we inspect the average inference time with each experiment's standard deviation (in seconds). Table \ref{tab:inference_time} reports that FlowMP achieves sub-$0.1$ seconds of mean inference time in \texttt{RobotPointMass} environment with all inspected trajectory lengths. Meanwhile, in \texttt{PandaSphere} environment, we achieve a slightly faster inference time than MPD diffusion prior inference. This experiment showcases that FlowMP is well-suited for real-time deployment with GPU-acceleration on a robot platform.

%% file: 05_experiments.tex
\subsection{Real-Robot Experiments}

We input $20$ collision-free, dynamically feasible trajectories with $L=256$ and state dimension $\bm{\xi} = \{q_i, \dot{q}_i, \ddot{q}_i\} \in \mathbb{R}^{21}$ to drive the Kinova Gen3 manipulator from the starting region to the goal region. FlowMP has a size of $4.5$ MB of $362,922$ trainble parameters. The inference runs on an Intel Core i9-14900 CPU and an NVIDIA RTX 4080S GPU, achieving an inference speed of approximately $0.1$ seconds.  
Fig.~\ref{fig:overview_method} presents the experimental results of executing a smooth, dynamically feasible motion on the Kinova Gen3 manipulator. Given the same start and goal configurations, FlowMP generates multiple valid solutions, demonstrating its ability to capture different modes of trajectory distribution. Furthermore, when varying the start and goal within a close region, the resulting motions remain smooth and dynamically feasible, making them directly practical to the robotic system, as shown in Fig. \ref{fig:overview_method}. The demonstration video of the experiments can be seen in the supplementary document.  

%% file: 06_conclusions.tex
\section{Conclusions}

We introduced FlowMP---a framework that leverages conditional flow matching to learn motion fields for generating smooth and dynamically feasible trajectories. Using flow matching to encode trajectory distribution and capture second-order trajectory dynamics, our approach directly models acceleration profiles alongside velocity, ensuring physically executable trajectories without requiring iterative denoising steps. Through evaluations, FlowMP demonstrated superior performance over MPD and bare trajectory optimizer in terms of trajectory quality, inference speed, and planning feasibility across both planar and robot environments. Our method not only achieves $\times 2$ faster inference times with scaling trajectory horizon but also maintains trajectory smoothness and scalability for varying trajectory lengths, addressing key limitations in prior diffusion-based approaches. Furthermore, real-robot experiments confirm FlowMP’s practical applicability in robotic manipulation tasks with very few or no blending steps, highlighting its potential for real-world deployment, even sampling directly from prior.